# RECOGNIZING THE VOCABULARY OF BRAZILIAN POPULAR NEWSPAPERS WITH A FREE-ACCESS COMPUTATIONAL DICTIONARY

Maria José Bocorny FINATTO*
Oto Araújo VALE**
Éric LAPORTE***

- ABSTRACT: We report an experiment to check the identification of a set of words in popular written Portuguese with two versions of a computational dictionary of Brazilian Portuguese, DELAF PB 2004 and DELAF PB 2015. This dictionary is freely available for use in linguistic analyses of Brazilian Portuguese and other researches, which justifies critical study. The vocabulary comes from the PorPopular corpus, made of popular newspapers *Diário Gaúcho* (DG) and *Massa*! (MA). From DG, we retained a set of texts with 984.465 words (tokens), published in 2008, with the spelling used before the Portuguese Language Orthographic Agreement adopted in 2009. From MA, we examined papers of 2012, 2014 e 2015, with 215.776 words (tokens), all with the new spelling. The checking involved: a) generating lists of words (types) occurring in DG and MA; b) comparing them with the entry lists of both versions of DELAF PB; c) assessing the coverage of this vocabulary; d) proposing ways of incorporating the items not covered. The results of the work show that an average of 19% of the types in DG were not found in DELAF PB 2004 or 2015. In MA, this average is 13%. Switching versions of the dictionary affected slightly the performance in recognizing the words.

- KEYWORDS: Popular newspapers. Lexis. Vocabulary. NLP dictionary. Lexical coverage. Word recognition. Brazilian Portuguese.

## Introduction

Written text of popular newspapers in Brazil has still been little investigated in language studies and is somewhat overlooked as a source of data to elaborate language resources for natural language processing (NLP), and even to feed the large corpora that represent current Portuguese. However, this type of source is considered to be a new

* Federal University of Rio Grande do Sul (UFRGS), Porto Alegre - RS – Brasil. PhD in Letters. maria.finatto@ufrgs.br. ORCID: 0000-0002-6022-8408

** Federal University of São Carlos (UFSCar), Center for Education and Human Sciences, São Carlos - SP - Brasil. Professor of Department of Letters. otovale@ufscar.br. ORCID: 0000-0002-0091-8079

*** Université Paris-Est (UPE), LIGM, UPEM/CNRS/ESIEE/ENPC, Champs-sur-Marne - France. Institut d'électronique et d'informatique Gaspard-Monge. eric.laporte@univ-paris-est.fr. ORCID: 0000-0002-0984-0781

and important type of vehicle for communication (AMARAL, 2006) and has already been used to generate data for some promising computational applications associated to the description of Brazilian Portuguese (BP) usage, as we see, for example, in Zilio (2015). The various Brazilian popular newspapers have also facilitated studies on new formats of journalistic genres in communication studies (TRISTÃO; MUSSE, 2013; SELIGMAN, 2008); among linguists, they have served as a source of teaching material for Portuguese as a native or foreign language (FINATTO; PEREIRA, 2014) or as objects of study in critical discourse analysis (SOARES, 2017).

The vocabulary of this new type of newspaper, however, challenges the coverage of NLP dictionaries. Also known as NLP lexicons, these dictionaries are designed with the particular purpose to be looked up by computer systems, functioning as parts of specific programs (see ZAVAGLIA, 2006). NLP dictionaries, called *léxicos computacionais* in Portuguese (CHISHMAN, 2016), are those dictionaries built specifically for use in NLP, and for their proper operation they require the careful incorporation of a whole range of linguistic descriptive data (cf. LAPORTE, 2013). Such incorporation involves a process of improvement throughout different versions of these dictionaries. Among different NLP dictionaries, we picked out DELAF PB, freely available for use in different research projects, in linguistics and in computer science, which justifies a critical study of its different versions.

Thus, having in mind the hypothesis that we might uncover gaps in existing dictionaries by using them to process the vocabulary of Brazilian popular newspapers, we carried out the experiment reported in this paper, which consisted in identifying the words[1] in this type of newspaper, and in observing if they are covered or recognised by the two versions of the free-access computational dictionary. The Brazilian popular newspapers that provided the vocabulary were *Diário Gaúcho* ("Daily of Rio Grande do Sul", henceforth DG) and Bahian newspaper *Massa!* ("Amazing!", MA). The texts we used can be consulted through search words on the website of the PorPopular Project.[2] More information on these media, their texts and more frequent vocabulary can be found in Amaral (2006) and Finatto (2012). Oliveira (2009) gives references on popular newspapers from other regions of Brazil.

As already mentioned, the language resource selected to test the recognition of the vocabulary of popular newspapers was DELAF PB, an NLP dictionary of BP distributed along with the UNITEX system (PAUMIER, 2016). We used two versions of this dictionary: DELAF PB 2004 (MUNIZ, 2004) and DELAF PB 2015 (VALE; BAPTISTA, 2015; CALCIA *et al.*, 2014). Although this dictionary is still little used by Brazilian linguists, there have been publications about it in Brazil for a long time (MUNIZ, 2004), as well as about its suitability for different research projects (e.g.,

---

1   The notion of word is quite controversial in the context of language studies, as taught by Biderman (1999, 1998). What a word is can be understood in several ways, hence the terms 'lexeme', 'lemma' or 'lexical unit/item/entry', 'vocable', among others, to denote different facets of this concept.

2   The texts that we used in this experiment can be consulted via the 'context generator' tool in: http://www.ufrgs.br/textecc/porlexbras/porpopular/experimente.php

ALMEIDA; FERREIRA, 2007; VICENTINI, 2010), both in NLP and in applied linguistics, which also justifies its examination in this work. Again, its access is free, and the software provides assistance for adaptation and insertion of entries according to the users' needs.

The DELAF PB dictionaries, thus, are built and updated collaboratively. They have been quite useful in different applications, especially NLP products. An example of a very recent and successful application in Brazil of the DELAF PB dictionaries is the work of Paiva, Barbosa, Faria and Martino (2017). The award-winning research of these authors brought together linguists and computer scientists and produced an automatic translator of written Brazilian Portuguese into the Brazilian Sign Language (LIBRAS).[3]

With the DELAF PB dictionary in its 2004 and 2015 versions, we also wanted to assess its update. Only the more recent version complies with the Portuguese Language Orthographic Agreement. The texts of the popular newspapers that we used were produced before (DG) and after (MA) the adoption of the new spelling. Thus, we could also observe the impact of the presence of two spelling standards on the performance of the resource.

In addition to the above, the general purposes of this work are:

a) to disseminate Brazilian popular newspapers as a source of research on the lexis in written form;
b) to contribute useful data for future extensions and improvements of DELAF PB.

The rest of this paper contains: a) data on the corpus of popular newspapers used in the experiment; b) a presentation of the dictionaries and a description of the operation of recognising or identifying the words in a corpus; c) the stages of the work, with comments and an overview of the results; d) a characterization of the majority profile of the items not covered by the dictionaries, and a discussion of how the performance of the dictionaries might be improved.

**Text corpus and samples under analysis**

From DG, we used as a starting point only part of the collection available in the PorPopular corpus. In the DG corpus, we selected texts published in 2008 that total 984,465 tokens. We obtained, therefore, a corpus with almost one million words, a significant size in this type of study. In this sample under examination to verify the vocabulary of DG, the texts are of varied types and correspond to what is published in the complete daily newspaper in its printed format. There are general news, horoscope columns, sports columns, police news and miscellaneous topics. Other parts of the DG corpus have already been used in previous studies (ZILIO, 2015; FINATTO *et al.*, 2011) for diverse purposes.

---

[3] This research and its result, which used the DELAF PB dictionary, were recognised as the best work in the Scientific Merit category, during the Meeting of Corpus Linguistics and Brazilian School of Computational Linguistics 2017 (http: // www.ufrgs.br/elc-ebralc2017).

For the corpus of the MA newspaper, we started from a smaller sample, much more specific in terms of text typology: a set of 724 texts, comprised only of news published in the online version of the newspaper and dealing with various themes.[4] These texts add up to 215,776 tokens and were published in 2012, 2014 and 2015.

The DG texts, as already mentioned, date from 2008, **before** the most recent orthographic change of BP. Information on the quantitative effect of this change on the spelling variation observed in newspapers can be read in Flores and Finatto (2009). For a general view of the impact of this orthographic alteration, see the collection organized by Moreira, Smith and Bocchese (2009).

This reform came in force before the MA texts were produced, but after the DG texts, so we had to deal with two spelling standards in our experiment. Thus, the words in the DG sample of 2008 still have hyphens, accents and tremas (ex: *agüentar*) which lack in the MA sample (ex: *aguentar*). These differences led us to use two versions of the DELAF PB dictionary, one for each spelling standard. Moreover, the use of both versions, from 2004 and 2015, also served to verify the comprehensiveness of the update, in terms of the coverage of the items used in these popular newspapers, with and without the new spelling.

Thus, we carried out the verification with both versions of the dictionary: DELAF PB 2004 (MUNIZ, 2004) and DELAF PB 2015 (VALE; BAPTISTA, 2015; CALCIA *et al.*, 2014).

**The DELAF dictionaries**

'DELAF' is, in fact, a format of NLP dictionaries that originated from research carried out for the French language by the team led by linguist Maurice Gross in the Laboratory of Documentary and Linguistic Automation (LADL).[5] This work was subsequently extended to other languages through the RELEX network of laboratories.[6]

DELAF dictionaries describe the simple words and multiword units of a given language by associating each form with both a lemma and a series of grammatical, semantic and inflectional codes.

The DELAF dictionaries have been developed by teams of linguists for various languages (French, English, Greek, Italian, Spanish, German, Thai, Korean, Polish, Norwegian, Portuguese, Arabic, among others) and are used today in academic and industrial projects. Several, among them the French and the Brazilian Portuguese DELAFs, are freely accessible and updated collaboratively.

---

[4] [1] This is 'Sample 2' of newspaper *Massa* in the PorPopular corpus. It is available for download in - http://www.ufrgs.br/textecc/porlexbras/porpopular/caixaferramentas.php#dadosCorpus

[5] For more information on LADL, see: http://infolingu.univ-mlv.fr/LADL/Historique.html

[6] Cf. http://unitexgramlab.org/pt/relex-network

**Principle of operation of the dictionaries in the UNITEX system**

UNITEX is a free-access corpus analyser that processes texts in natural languages with the aid of language resources. These resources are NLP dictionaries in the DELAF format, grammars, and Lexicon-Grammar tables integrated into the system. Some language resources are distributed along with UNITEX and others can be elaborated by users. The UNITEX system is freely accessible and currently used to support studies on different languages (PAJIĆ et al., 2018), including ancient languages (KINDT, 2018).

As shown by Almeida and Ferreira (2007), the UNITEX system allows for processing any set of texts, so that the linguistic expressions in it are located and categorized. Locating or identifying expressions in a given text or corpus will be effective "provided that such expressions are represented in the dictionary coupled with it" or that the user describes them in a query. UNITEX allows for identifying words by classes.

Once some text has been selected, such as that of our popular newspapers, UNITEX offers to preprocess it. Such preprocessing consists in applying to the text the following operations: normalizing delimiters, segmenting into tokens, normalizing unambiguous forms, segmenting into sentences and, finally, applying NLP dictionaries present in the computer. The presence of these dictionaries in the UNITEX system is a differential with other usual search tools for finding word patterns in corpora, since one can find large classes of words with simple patterns.

When a given corpus or text in a given language is processed, the internal operation of UNITEX consists of constructing a subset of the dictionaries, with only the forms present in the corpus being processed. Thus, for example, the application of the DELAF dictionary on a text such as *O time de Neymar corria atrás do prejuízo* 'Neymar's team was trying hard to make up lost ground' [lit. 'Neymar's team was running after the loss'] will produce the following subset of the dictionary of simple words:

```
atrás,.ADV
corria,correr.V:I1s
corria,correr.V:I3s
de,.PREP
do,.PREPXDET+Art+Def:ms
do,.PREPXPRO+Dem:ms
o,.DET+Art+Def:ms
o,.N:ms
o,.PRO+Dem:ms
o,ele.PRO+Pes:A3ms
prejuízo,.N:ms
time,.N:ms
```

The name *Neymar*, being described in a dictionary of proper names distinct from the Portuguese DELAF, will be considered an 'unknown word'.

The application of the dictionaries to the text is performed by the UNITEX system with its program called DICO and generates subsets — 'sub dictionaries' — called *simple words; compound words; unknown words*. In this work, we only deal with the last group.

UNITEX has a second use: it offers support for easily inserting new words and grammatical information into the user's NLP dictionaries, thus adapting them to different purposes. The UNITEX dictionaries deploys the formalism of the DELA (Electronic Dictionaries of the LADL). This formalism allows for describing the simple and multiword lexical entries of a language, optionally assigning them grammatical, semantic and inflectional information. Within this formalism, two kinds of dictionaries are distinguished. The most frequently used kind is the dictionary of inflected forms, in the DELAF format (DELA of inFlected forms). The second kind is the dictionary of lemmas, in the DELAS (DELA of simple words) or DELAC (DELA of compound words) formats, which generates the other dictionaries. In this work, we deal only with DELAF resources. An entry of DELAF PB is organized as follows:

sambou,sambar.V:J3s

Here, *sambou* 'danced samba' is the form found in the text, 'sambar' the lemma, 'V' the part of speech — in this case a verb — and 'J3s' the inflectional code — in this case a third person singular preterite. The complete list of grammatical and inflectional codes for BP can be found in Muniz (2004).

For BP, the UNITEX system comes with an NLP dictionary in the following two versions:

- the 2005 version: from a DELAS with 61,335 words, a DELAF with 878,095 simple inflected words and 4,100 multiword inflected units was generated. These resources were created from the dictionary of ReGra — the base of the BP spell checker of Word for Windows — for nouns, adjectives and adverbs (MARTINS *et al.*, 1998), and from Vale (1990) 102 verbal inflection templates.
- the 2015 version: the forms with the new spelling resulting from the 1990 Orthographic Agreement were incorporated. 7,900 new entries of simple lemmas (nouns, adjectives and adverbs) were introduced, in addition to verb forms with enclitics and mesoclitics, which were not in the first version. With these changes, the DELAF 2015 of Brazilian Portuguese now totals 10,954,724 entries, describing 7,632,498 unique forms.

**Stages of work and results**

In general terms, the verification experiment involved:

a) generating the list of words (types) used in the DG newspaper—without changing the spelling;

b) generating the word list of the MA newspaper;

c) comparing each list with the entry lists of the 2004 and 2015 versions of the NLP dictionary DELAF PB;

d) assessing the lexical coverage of each sample — in terms of tokens and types — by DELAF PB, in both versions;

e) proposing ways of including items not identified by the dictionaries.

In these stages, the verification generated two lists of words unknown by the DELAF-PB from each newspaper, for each of the two spelling standards (lists $DG_{04}$, $DG_{15}$, $MA_{04}$ and $MA_{15}$). Then case sensitivity was removed.

At the beginning of the comparison process, we realized DELAF PB 2015 did not contain the list of abbreviations and acronyms—respectively labelled ABREV and SIGL — which were in the 2004 version of the dictionary. As a matter of fact, during the revision by Calcia *et al*. (2014), the forms of abbreviations and acronyms (such as *ABS* or *ABNT* in the DG newspaper, or *UFBA* or *UFC* in MA) were not the object of study in the update of the dictionary and were moved into a separate dictionary. To standardize the experimental conditions, lists $DG_{15}$ and $MA_{15}$ were generated again using DELAF PB 2015 along with the dictionary of abbreviations and acronyms. In what follows, the statistics for DELAF PB 2015 are the result of this second generation.

**Tables 1 and 2** below reproduce parts of each of these lists. We display the first 30 items unknown by the UNITEX dictionaries starting with the letter U, and then the first 30 with initial A:

**Table 1** – Sample of list of items from DG and MA
unknown in DELAF PB, starting with U.

| Items starting with U | DG₀₄ DELAF 2004 | DG₁₅ DELAF 2015 | MA₀₄ DELAF 2004 | MA₁₅ DELAF 2015 |
|---|---|---|---|---|
| 1. | uai | uai | ualex | ualex |
| 2. | uau | uau | uanderson | uanderson |
| 3. | ubial | ubial | ubandista | ubandista |
| 4. | ubs | ubs | ubang | ubang |
| 5. | udesca | udesca | ubatã | ubatã |
| 6. | udi | udi | ubiracê | ubiracê |
| 7. | údiche | údiche | ucla | ucla |
| 8. | udine | udine | uefs | uefs |
| 9. | udinese | udinese | uellinton | uellinton |
| 10. | uebel | uebel | uelliton | uelliton |
| 11. | uefa | uefa | uenf | uenf |
| 12. | ufa | ufa | ueslei | ueslei |
| 13. | ufcspa | ufcspa | uezo | uezo |
| 14. | uflacker | uflacker | ufc | ufc |
| 15. | ufsm | ufsm | uff | uff |
| 16. | ugapoci | ugapoci | ufrrj | ufrrj |
| 17. | ughini | ughini | uhu | uhu |
| 18. | ugowski | ugowski | uibaí | uibaí |
| 19. | uilson | uilson | ulício | ulício |
| 20. | ulalá | ulalá | umidificador | unasul |
| 21. | ulbra | ulbra | unasul | under |
| 22. | uli | uli | under | undime |
| 23. | ulmen | ulmen | undime | uneb |
| 24. | ulsan | ulsan | uneb | unifacs |
| 25. | ultramen | ultramen | unifacs | unifcas |
| 26. | ultrasom | ultrasom | unifcas | unirio |
| 27. | ultrassonografias | umbom | unirio | unit |
| 28. | umbom | umchorão | unit | united |
| 29. | umchorão | umespa | united | universitario |
| 30. | umespa | unasul | universitario | uol |

**Source**: Author's elaboration.

**Table 2** – Sample of list of items from DG and MA
unknown in DELAF PB, starting with A.

| Items starting with A | $DG_{04}$ DELAF 2004 | $DG_{15}$ DELAF 2015 | $MA_{04}$ DELAF 2004 | $MA_{15}$ DELAF 2015 |
|---|---|---|---|---|
| 1. | aabb | aabb | abadá | abadá |
| 2. | Aaliyah | aaliyah | abadábraço | abadábraço |
| 3. | aas | aas | abadás | abadás |
| 4. | abachilov | abachilov | abaralhau | abaralhau |
| 5. | abadía | abadía | abdelmassih | abdelmassih |
| 6. | abandon | abandon | abdulá | abdulá |
| 7. | Abatê | abbey | abefin | abefin |
| 8. | abbey | abbott | aberbach | aberbach |
| 9. | abbott | abdel | abisson | abisson |
| 10. | abdel | abdômem | abla | abla |
| 11. | abdômem | abdul | abordá | aboubacar |
| 12. | abdominoplastia | abdulla | aboubacar | abravanel |
| 13. | Abdul | abebe | abravanel | academiagf |
| 14. | abdulla | abech | academiagf | accosta |
| 15. | Abebe | abelão | accosta | acessando |
| 16. | Abech | abelhocídio | acessando | acessar |
| 17. | Abelão | abenício | acessar | acesse |
| 18. | abelhocídio | ablo | acesse | acm |
| 19. | abenício | about | acm | adab |
| 20. | Ablo | abp | adab | adailson |
| 21. | abordá | abração | adailson | adailton |
| 22. | aborígenes | abraciclo | adailton | adan |
| 23. | About | abramet | adan | adanascimento |
| 24. | Abp | abramovich | adanascimento | adecir |
| 25. | Abraçá | abrh | adecir | adelmário |
| 26. | abração | abrhrs | adelmário | adelmo |
| 27. | abraciclo | abrigagem | adelmo | ademi |
| 28. | abramet | abrilina | ademi | ademilson |
| 29. | abramovich | abrito | ademilson | adenilton |
| 30. | Abrh | abs | adenilton | aderam |

**Source**: Author's elaboration.

After the generation of lists of unknown words by the DELAF 2004 and 2015 dictionaries, these lists were compared to one another, ignoring case. The items were studied as to their use in texts and considering the information recorded in two conventional dictionaries of BP: Aurélio (FERREIRA, 1999) and Houaiss (HOUAISS, VILLAR, 2009). They were tentatively grouped into categories such as:

(1) **Typing errors** (*umchorão, ubandista*);

(2) **Old spellings** (*idéia* 'idea');

(3) **Proper names** (*uilson, uanderson*);

(4) **Abbreviations/acronyms** (*abs, ufrrj*);

(5) **Diverse expressions/slang/foreignisms** (*ulalá* 'my god!'*, university, united*);

(6) **Other nouns** (*umidificador* 'humidifier');

(7) **Other** (*abadábraço, aboubacar*).

These categories, of course, were a tentative first approach to the out-of-coverage items in the lists, and they can be refined in future work. Some words can be classified as neologisms or regionalisms, for example. Some are at the same time a noun and a neologism (cf. *abelhocídio* 'bee killing'). We found virtually no adjectives or verbs among out-of-coverage items, so we did not establish categories 'verb' or 'adjective' in this initial approximation. A multifactorial categorization of out-of-coverage items would mean another work.

**Results: overview and summary**

We summarize in Table 3 below the main results obtained from the two samples of popular newspapers. These results will be discussed in the next section.

**Table 3** – Results obtained from DG and MA.

| Newspaper | *Diário Gaúcho* (DG) – sample of diverse texts | *MASSA!* (MA) – sample of news |
|---|---|---|
| Spelling | old | Present |
| Types | 53.966 | 22.414 |
| Tokens | 984.465 | 215.776 |
| **With DELAF 2004:** | | |
| out-of-coverage types | 10.512 | 3.048 |
| % of types | 19,48% | 13,60% |
| out-of-coverage tokens | 36.190 | 11.624 |
| % of tokens | 3,68% | 5,39% |
| **With DELAF 2015:** | | |
| out-of-coverage types | 9.967 | 2.769 |
| % of types | 18,47% | 12,35% |
| out-of-coverage tokens | 34.611 | 10.870 |
| % of tokens | 3,52% | 5,04% |

**Source**: Author's elaboration.

### Considerations on the results - Identification of the vocabulary

DELAF PB has a large lexical coverage of twentieth-century newspapers and nineteenth-century literary texts (1.9% of types are out of coverage in the novel *Senhora* by José de Alencar). By comparison, the percentages of out-of-coverage types in Table 3 (from 12% to 19%) are appreciably higher. Therefore, the vocabulary of the kind of newspaper under study can be seen as an important impediment to the identification of words by the DELAF-PB dictionaries. This is relevant for BP researchers interested in its future use.

In order to contextualize the results summarized in the previous section, it is important to remember the factors involved in the identification of items in the vocabulary of popular newspapers by the DELAF PB dictionaries:

a) DG texts have words in the old spelling standard—before the agreement;

b) MA texts have words in the present spelling;

c) only DELAF PB 2015 complies with the new spelling;

d) DELAF PB 2004 does not include the new spelling.

The list of words (types) employed in DG, a set of 53,966 items, includes 19.48% of items unknown by DELAF 2004 and 18.47% of items not covered by DELAF 2015. Thus, there is a small reduction in this percentage of **out-of-coverage words** between the two versions of the dictionary.

The DG corpus does use an old spelling standard, but the fact that old spellings, such as *agüentar*, are not present in DELAF 2015 seems to have affected the performance less than the insertion of new words, such as *umidificador* ('humidifier'), and of verb forms with clitics, such as *abordá-lo* ('approach him/it'). So, the percentage of out-of-coverage words has decreased.

In the MA newspaper, we had 22,414 types, of which 13.60% are unknown in DELAF 2004 versus 12.35% in DELAF 2015. In this case, the spelling is entirely in the present standard. Again, text coverage by DELAF improved from 2004 to 2015. This time, the effect of the adaptation of the dictionary to the spelling agreement added to the effect of inserting new words and verbs with clitics.

Thus, both in the MA corpus and in the DG corpus, from DELAF 2004 to DELAF 2015, the performance improved slightly with respect to the recognition of items from the popular newspaper. The recognition of items, from 2004 to 2015, increased, on average, by 1.13 percentage point in terms of recognised types.

In terms of number of tokens, the DG vocabulary is noticeably more covered (96.4% on average) than the MA vocabulary (on average 94.8%), with little effect of the update of the dictionary on this performance. Several kinds of reasons, of course, may explain this. However, it is worth considering that the MA newspaper is from the Northeast region of Brazil, and DG from the South region, which can affect the observed lexical profile. In terms of types, the difference in coverage appears to be the other way around, but this comparison is not significant because of the difference in sample size: in a larger sample, such as that of MA, statistics on the number of types lead to over-representation of infrequent words, which are less likely to appear in the dictionary.

In any case, popular newspapers prove to be an interesting and challenging object of research. However, different current corpora of Brazilian Portuguese are still composed only, as a rule, of data collected from large traditional newspapers, e.g. *Folha de S. Paulo* (ALUISIO; ALMEIDA, 2006).

As we have seen, for the DELAF PB dictionaries, the vocabulary of popular newspapers seems somewhat "odd". Thus, future work could contrast the percentage of out-of-coverage words between popular and traditional newspapers from the same period.

Another interesting question about out-of-coverage vocabulary is its potential connection with regionalist or local topics (such as the noun *cacetinho* in DG, the equivalent of *pão francês*, 'smaller French bread', in Southern Brazil) or with newly created proper names (such as *Abadábraço* in MA).

In the next section, we outline the most frequent categories of words unknown by the two versions of the dictionaries, based on our samples. Then we briefly discuss how some out-of-coverage items could be incorporated into the DELAF PB 2015 dictionary.

**Profile of out-of-coverage words and options for extension of DELAF PB 2015**

As we have already noted, the examination of the lists of words in our tables shows that the 2015 version of DELAF PB achieved only a modest gain in coverage for this particular type of newspaper. Thus, the listing of the first 30 words that start with A is practically identical in both columns that refer to the MA corpus. We note the presence of regional vocabulary (*abadá* 'sleeveless shirt', *abaralhau* 'a Bahian dish with peeled beans and salt cod' - MA) and a set of proper names. Although DELAF PB contains a good number of proper names (*Aldemário*, *Abramovich*) that would not normally appear in a conventional dictionary, there is still good work to be done in listing and characterizing proper names, in particular because of their rich variation in Brazil (see *Uanderson, Uellinton, Ueslei*).

A closer examination of the lists also shows that most of the unrecognized items are nouns (in DG: *abrigagem* 'sheltering', *abdômem* 'abdomen', *abelhocídio* 'bee killing'), including proper names (*Abelão*). The only two verbs identified in the lists presented here were the verb *acessar* 'access', with several forms, and the form *abordá* of the verb *abordar* 'approach', one of the forms with clitics, which were not covered in the 2004 version of DELAF PB (*abordá-lo* 'approach him/it').

Abbreviations and acronyms are another issue. Mastering the construction of NLP dictionaries requires a special effort, integrating linguists and computer scientists. The construction of more comprehensive computational resources, taking into consideration recurrent processes and phenomena of the current Portuguese lexicon, is a very complex challenge. In particular, abbreviations, acronyms and proper names are recurrent phenomena in written language and require specific work to preserve and improve the operation of NLP systems. Papers such as Vale *et al*. (2008) have already pointed out this need in the case of Portuguese historical corpora, also encompassing contemporaneous text collections. As a matter of fact, building specific NLP dictionaries for abbreviations, acronyms and named entities could be an appropriate way to address the challenge we posed to the UNITEX system with our popular newspapers. Of course, as we can see in the following excerpts from two news stories in our corpus, there is much more to be explored:

EXCERPT 1:

*A primeira delas é o lançamento do Abadábraço, um bloco que desfilará sem cordas, mas com os foliões. Desfilará sem cordas, mas com os foliões devidamente trajados com abadás. A proposta aqui é incluir. É ter mais pessoas brincando nas ruas e com direito a usar o seu abadá.* ('The first of them is the launch of Abadábraço, a Carnival group that will march without a rope line, but with its members.[7] It will parade without a rope line, but the revellers will wear the proper shirts. The proposal here is to be inclusive. It's to have more people having fun in the streets and being allowed to use their shirt.')

EXCERPT 2:

*Chaleira, César Oliveira & Rogério Melo, Bochincho, Os Quatro Gaudérios, Portal Gaúcho e Eco do Minuano & Bonitinho. Foi grande a integração entre as invernadas adulta e xiru na Sociedade Gaúcha de Lomba Grande, em Novo Hamburgo, que comemorou 70 anos na noite de terça-feira.* ('Chaleira, César Oliveira & Rogério Melo, Bochincho, The Four Wanderers, Portal Gaúcho and Eco do Minuano & Bonitinho. There was a strong sense of togetherness between the adult and senior dance groups at the Club of Rio Grande do Sul at Lomba Grande, in Novo Hamburgo, which celebrated its 70th birthday on Tuesday night.')

**Conclusion**

The research problem addressed in this work was to describe and discuss the performance of a large-coverage NLP dictionary, freely available to be used for research on Brazilian Portuguese and in NLP industry.

The DELAF dictionary under examination, although a quite valuable help for different linguistic tasks, might be extended with data from corpora of Brazilian popular newspapers. After all, as already mentioned, this journalistic genre has still been little contemplated as a source of data for the study of cultured written Portuguese. Such lexical incompleteness, in terms of number of out-of-coverage tokens, is noticeably lower in the vocabulary of the newspaper from Rio Grande do Sul (on average, 96.4%) than from Bahia (an average of 94.8%), a fact that motivates further research. Therefore, by examining the performance of different versions of the dictionary in processing the vocabulary of Brazilian popular newspapers, we have demonstrated the validity of taking both types of resources as objects of study and as sources of data for language research conducted in cooperation by linguists and computer scientists.

---

[7] In the Bahian Carnival, each marching group of revellers is traditionally surrounded by a rope line, and recognisable by the distinctive sleeveless shirt worn by its members. The name *Abadábraço* incorporates the words for 'hug' and for this type of shirt (TN).

**Acknowledgments**

This study was financed in part by the Coordenação de Aperfeiçoamento do Pessoal de Nível Superior – Brasil (CAPES) in the scope of the CAPES-STIC-AMSud (proj.047/2014), by the the CNPq – National Council for Scientific and Technological Development (PQ Grant – proc. 305625/2016-0 and APV Grant proc. 453058/2015-9), and Fundação de Amparo à Pesquisa do Estado de São Paulo – FAPESP (proc. 2016/24670-3).

FINATTO, M.; VALE, O.; LAPORTE, E. Reconhecimento do vocabulário de jornais populares brasileiros por um dicionário computacional de acesso livre. **Alfa**, São Paulo, v.63, n.1, p.67-85, 2019.

- *RESUMO: Relata-se um experimento de verificação da identificação de um universo de palavras do português popular escrito por duas versões de um dicionário computacional do português brasileiro (PB), DELAF PB 2004 e DELAF PB 2015. Esse dicionário computacional é gratuitamente acessível para ser utilizado em análises linguísticas do Português do Brasil e em outras pesquisas, o que justifica um estudo crítico. O universo vocabular provém do corpus PorPopular, composto por jornais populares, o Diário Gaúcho (DG) e o jornal baiano Massa! (MA). Do DG, partiu-se de um conjunto de textos com 984.465 palavras (tokens), publicados em 2008, com ortografia desatualizada frente ao Acordo Ortográfico da Língua Portuguesa adotado em 2009. Do MA, examinou-se um universo com 215.776 palavras (tokens), em publicações de 2012, 2014 e 2015, com todo o material na nova ortografia. A verificação envolveu: a) gerar listas de palavras diferentes empregadas em DG e MA; b) comparar essas listas com as listas de entradas das duas versões do DELAF PB; c) avaliar a cobertura desse vocabulário; d) propor modos de inclusão de itens não cobertos. Os resultados do trabalho mostraram, no DG, uma média de 19% de palavras diferentes (types) desconhecidas pelos DELAF PB 2004 e 2015. No MA, essa média ficou em 13%. A versão do dicionário repercutiu ligeiramente sobre o desempenho do reconhecimento de itens.*

- *PALAVRAS-CHAVE: Jornais populares. Léxico. Vocabulário. Dicionário computacional. Cobertura lexical. Reconhecimento de palavras. Português brasileiro.*